\documentclass[10pt,twocolumn,letterpaper]{article}

\usepackage{cvpr}              % To produce the CAMERA-READY version

\usepackage{lipsum}
\usepackage{multirow}
\usepackage{tabularx}
\usepackage{makecell}
\usepackage{amsmath}
\usepackage{color, colortbl}
\usepackage{nccmath}
\usepackage{float}
\usepackage{afterpage}
\usepackage[accsupp]{axessibility} % Improves PDF readability for those with visual impairments.
\usepackage[normalem]{ulem}
\usepackage[dvipsnames]{xcolor}

\definecolor{cvprblue}{rgb}{0.21,0.49,0.74}
\usepackage[pagebackref,breaklinks,colorlinks,citecolor=cvprblue]{hyperref}
\definecolor{tablegray}{rgb}{0.9,0.9,0.9}

\def\paperID{6191} % *** Enter the Paper ID here
\def\confName{CVPR}
\def\confYear{2024}

\title{Multi-agent Long-term 3D Human Pose Forecasting via Interaction-aware Trajectory Conditioning}

\author{Jaewoo Jeong$^{*}$, Daehee Park$^{*}$, and Kuk-Jin Yoon\\
KAIST\\
{\tt\small \{jeong207,bag2824,kjyoon\}@kaist.ac.kr}}

\begin{document}
\maketitle
\def\thefootnote{*}\footnotetext{Denotes equal contribution.}\def\thefootnote{\arabic{footnote}}
\begin{abstract}
Human pose forecasting garners attention for its diverse applications.
However, challenges in modeling the multi-modal nature of human motion and intricate interactions among agents persist, particularly with longer timescales and more agents.
In this paper, we propose an interaction-aware trajectory-conditioned long-term multi-agent human pose forecasting model, utilizing a coarse-to-fine prediction approach: multi-modal global trajectories are initially forecasted, followed by respective local pose forecasts conditioned on each mode.
In doing so, our \textbf{T}rajectory\textbf{2P}ose model introduces a graph-based agent-wise interaction module for a reciprocal forecast of local motion-conditioned global trajectory and trajectory-conditioned local pose.
Our model effectively handles the multi-modality of human motion and the complexity of long-term multi-agent interactions, improving performance in complex environments.
Furthermore, we address the lack of long-term (6s+) multi-agent (5+) datasets by constructing a new dataset from real-world images and 2D annotations, enabling a comprehensive evaluation of our proposed model.
State-of-the-art prediction performance on both complex and simpler datasets confirms the generalized effectiveness of our method.
The code is available at \url{https://github.com/Jaewoo97/T2P}.
\end{abstract}    
\section{Introduction}
\label{sec:intro}

% Paragraph 1.
% Human motion forecasting 설명

% Human motion forecasting 은 과거의 사람의 physical motion 을 보고 미래 모션을 예측하는 것이다.
% 이는 human intelligence가 본능적으로 수행할 수 있는 능력으로, 이로 인해 사람은 군중 속에서도 자연스럽게 navigate 하거나, 가능한 위험을 파악할 수 있게 된다. 
% 이러한 이유로 human motion forecasting 은 자율주행과 surveillance 과 같은 다양한 computer vision task 에 중요한 역할을 하고 있다. 

% ChatGPT 버전
Human pose forecasting aims to predict future human motion based on observed past motion~\cite{guo2022multi, salzmann2022motron, mao2022weakly, ma2022progressively, maeda2022motionaug, zhong2022spatio, mao2021generating}.
Humans instinctively perform such tasks, allowing them to naturally navigate in crowded areas or identify and circumvent potential dangers.
For this reason, human pose forecasting plays an important role in various computer vision tasks~\cite{ham2023cipf,zheng2022multi,salzmann2023robots,huang2023diffusion,zhuo2019unsupervised, kim2023addressing}.
% such as autonomous driving~\cite{ham2023cipf,zheng2022multi}, robot planning~\cite{salzmann2023robots,huang2023diffusion}, and surveillance~\cite{zhuo2019unsupervised, kim2023addressing}.
Indeed, recent years have seen a proliferation of work on multi-agent motion forecasting which aim towards modeling complex multi-agent interaction~\cite{xu2023joint,guo2022multi,salzmann2022motron,peng2023trajectory,Xing_2023_ICCV}.

% Paragraph 2.
% - multi-agent motion forecasting 지금까지 방식의 한계점
%   - 기존 연구의 한계점:
%       - multi-agent간의 interaction을 고려하지만, 실제 환경에 맞지 않게 2-3명 기반의 데이터셋 밖에 없음
%           - JRT 같은 경우에는 attention이 비효율적으로 연산되어 memory 사용량이 큼, 10+ 명의 motion을 예측할 때 리소스 사용량이 매우 큼.
%       - Short-term 예측을 위한 attention network이기에 long-term 예측 성능이 좋지 않음. 
% - long-term, multi-agent 예측의 중요성 설명

% 이에 따라 다양한 기존 방법들이 제안되었으나, 기존 방법들은 크게 두가지 한계점을 가지고 있다.
% 첫번째는 long-term 예측을 수행하지 못한다는 것이다. 기존 연구들은 길어야 2초 정도를 예측하였다. 하지만 자율주행과 같은 다양한 downstream task 에 유용하려면, 예측 horizon 이 잠재적 위험을 파악할 수 있을만큼 충분히 길어야 한다. 
% 두번째는 multi-person 간의 interaction 이 제대로 학습되고 있지 않다는 점이다. 기존의 방법은 여러 사람의 joint 를 모두 한번에 interaction 의 고려 대상으로 생각한다. 하지만 이런 방법은 비효율적이며, 후에 실험적으로 보여주겠지만 interaction 을 motion prediction 에 제대로 활용 할 수 없도록 방해한다. 
Although various methods have been proposed,
% to fulfill these diverse needs
they share two major limitations.
The first is a limitation on long-term predictions, as previous studies predicted up to 3 seconds at most~\cite{peng2023trajectory,xu2023joint,xu2023stochastic,Barquero_2023_ICCV}.
%and performance starkly degrades beyond these timeframes.
However, a sufficiently long forecast horizon is essential to fully leverage human pose forecasting for diverse downstream tasks in the scope of identifying potential danger or understanding human behavior.
% such as autonomous driving, 
%for the observer to anticipate and plan interaction with the subjects.
The second is that multi-person interactions are not proficiently learned. 
Existing methods consider the joints of multiple people all at once as objects of interaction~\cite{peng2023trajectory,xu2023joint,vendrow2022somoformer}, resulting in an excessive complexity with respect to the number of joints. 
Due to such inefficient modeling, these approaches are found to be incompetent in long-term (3s+) multi-agent (6+) settings, limiting their practicality on complex real-world environments.
% These limitations aggravate 

\begin{figure}[t!]
\includegraphics[width=\columnwidth]{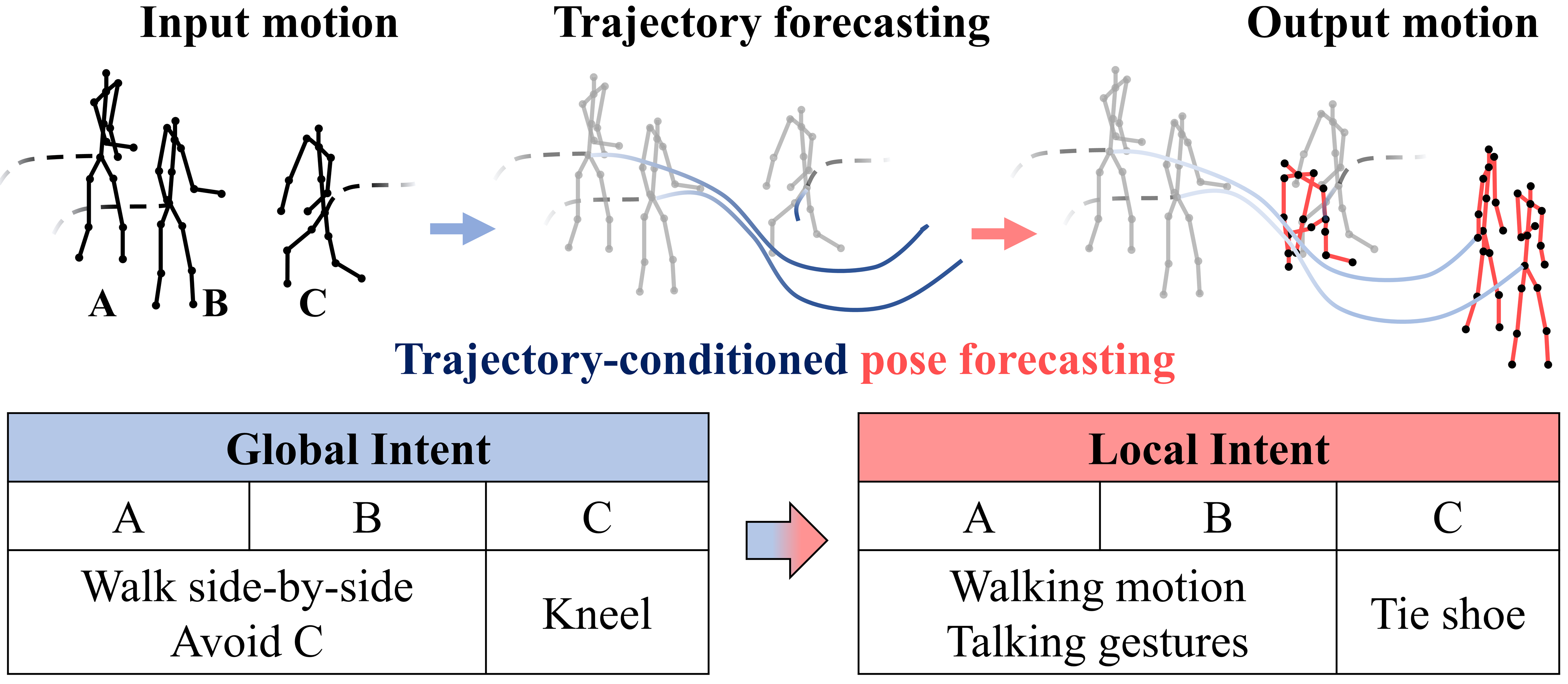}
\vspace{-17pt}
\caption{
% Since global intention contains hints for local intention. 
Human motion is goal-directed and influenced by other entities.
Therefore, global intention contains hints for local intention, allowing us to infer local pose from global trajectories.
%of agents.
% Indeed, we can infer local pose from global trajectories of agents.
Our method first forecasts global trajectories, upon which local poses are conditioned for subsequent forecasts. 
Pose and trajectory-wise inter-agent interactions are considered for both predictions.
% Both are predicted in a reciprocal manner while considering inter-agent interactions.
}
\vspace{-5pt}
\label{fig1_abstractFigure}
\end{figure}

% Paragraph 3.
% - Long-term, multi-agent 예측을 하는 데이터셋의 부재
% - 기존의 데이터셋은 raw agent의 수가 max 2~3명임, 이걸 섞어서 6~10명까지 있는 데이터셋을 synthetic하게 만듬.
% - 또한, 3dpw를 제외하곤 lab 환경에서 수집된 인위적인 human motion임. Real world 환경에서 long-term, multi-agent interaction을 담은 데이터셋이 없음.

% 더욱이, 이러한 두가지 한계점은 기존 데이터셋의 한계에도 그 원인이 있다. 
%기존의 존재하는 motion prediction 데이터셋은 sequence의 길이가 제한적이다. 또한 대부분의 데이터셋이 3명 이하의 사람이 존재하는 motion 을 포함하고 있다. 
% 일부 논문에서 여러 scene에 따로따로 존재하는 사람을 임의로 한 scene 에 더해서 최대 10명 까지의 사람이 존재하는 데이터셋을 만들긴 하였다. 
% 하지만 이 데이터같은 경우는 다른 scene 에서 온 사람들은 서로의 영향을 전혀 받지 않기 때문에, 의미있는 multi-person dataset 이라고 보기 어렵다. 
% 이러한 데이터셋의 한계로 인해 long-term, 그리고 multi-agent 환경에서 동작하는 model 이 개발될 기회가 없었다.
Moreover, these challenges are also due to the limitations of datasets. 
Existing pose forecasting datasets have limited sequence length ($\sim$3s) and number of agents ($\sim$2). 
Therefore, previous works~\cite{wang2021multi,peng2023trajectory,xu2023stochastic} have randomly blended disparate datasets to model multi-agent interaction with up to 10 agents.
% Yet, such naively merged data contains limited authentic interaction since agents from different scenes are not influenced by each other at all.
Yet, such naively merged data lacks authentic interaction as agents from different scenes remain uninfluenced.
As such, there was no opportunity to develop and evaluate a model in a long-term multi-agent environment.

% Paragraph 4. 
% - t2p 방법론 설명
%   - Trajectory encoding & local motion encoding 기반 trajectory prediction
%   - Graph-based agent interaction modeling
%   - Trajectory prediction 의 multi-mode proposal feature 기반 local motion generation
%   - Efficient decoupling of human motion into global & local motion
% - 데이터셋 설명
%   - 어떻게 만들었는지

% Long-term multi-agent human pose forecasting 을 가능하게 하기 위해, 우리는 모델과 데이터셋 두가지 관점에서 해결법을 제시한다.
% 우선 모델 관점에서, 우리는 Global trajectory-conditioned motion prediction 방법을 제안한다.
% 우리는 joint 단위로 모두 interaction 과 prediction 을 수행하는 기존 방법의 비효율성이 성능 저하의 원인이라고 지적한다.
% 이 비효율성을 해결하기 위해, 우리는 coarse-to-fine manner 의 방식을 제안한다. 
% 바로, global trajectory 를 활용하는 것이다. 
% global trajectory 는 사람의 hip joint 의 global 좌표에서의 움직임이고, 이는 사람의 대략적인 움직임을 의미한다.
% 사람의 대략적인 움직임, 즉 global trajectory 에는 여러가지 힌트가 담겨 있다.
% 예를 들어, 그림 1 에서 보면 두 사람이 나란히 걸어가고 있는 경우에는 두 사람이 상호작용하며 계속 걸어가는 중이라고 있는 중이라고 추측할 수 있다. 
% 두 사람이 접근하는 경우에는 그 둘이 악수를 하거나 인사 motion 을 취할 것이라고 추측할 수도 있다. 
% 또는 정지한 global trajectory 의 경우에는 그 사람이 가만히 앉거나 서있을 것이므로, 하체 joint 의 모션이 적을 것이라고 추측할 수 있다. 
% 이러한 사람의 대체적인 motion 은 joint 를 한번에 모두 예측하는 것보다 intention, interaction 을 파악하고 학습하기 쉽다는 것이 기존 연구에서 알려져 있다.
% 따라서 우리는 사람의 global trajectory 를 먼저 예측하고, 그에 conditioned 되어 joint 를 예측함으로써 미래 예측 task 의 multi-modal 한 특성을 반영한 모델 구조를 제안한다.
To this end, we present a solution from both model and dataset perspectives to tackle long-term multi-agent human pose forecasting.
First, from a model perspective, we propose an interaction-aware trajectory-conditioned pose forecasting method.
We point out that the limitations of existing methods on long-term multi-agent environments lead to poor performance in handling the multi-modal nature of human motion and correspondingly complex interactions.
% Excessive complexity of a joint-wise interaction modeling approach causes such joint ineffectiveness.
% We point out that existing methods are ineffective in forecasting handling the multi-modality of human motion with a joint-wise interaction modeling approach due to its excessive complexity~\cite{wang2021multi,peng2023trajectory,wang2021multi}.
% We point out that the inefficiency of existing methods that perform interaction and prediction on a joint basis is the cause of poor performance.
% To improve upon handling multi-modality in these complex settings for proficient motion forecasts, we use a coarse-to-fine approach.
To improve upon handling multi-modality in these complex settings, we use a coarse-to-fine approach to enjoy effective interaction modeling by propagating agent-wise coarse representations.
Agent-wise pose and trajectory embeddings are obtained in their respective local coordinates, followed by a holistic interaction modeling via our proposed Traj-pose module.
Interaction-aware forecasts are then made by initial \textbf{coarse} global hip joint trajectory forecast followed by \textbf{fine} local pose forecasts in its hip joint coordinates, conditioned on the global trajectory as shown in Fig.~\ref{fig1_abstractFigure}.
% Namely, initial \textbf{coarse} global hip joint trajectory forecast followed by \textbf{fine} local pose forecasts in its hip joint coordinates, conditioned on the global trajectory as shown in Fig.~\ref{fig1_abstractFigure}.
% As discovered in previous research~\cite{adeli2021tripod, salzmann2023robots}, learning the global intent is less challenging than predicting every joint-wise motion.
% Modeling agent interaction in such coarse latent space is found to be effective in handling complex multi-modality.
As discovered in previous research~\cite{adeli2021tripod, salzmann2023robots}, learning an agent-wise global intention as coarse trajectories is less challenging than predicting every joint-wise motion.
We leverage these hints from global trajectories, which are further conditioned towards forecasting local motion that embodies the interaction-aware spatio-temporal context.

From a dataset perspective, we parsed a novel real-world dataset for long-term multi-agent human pose forecasting.
We utilize JRDB dataset~\cite{vendrow2023jrdb} which consists of multi-view video and collected in various environments.
% JRDB dataset is collected by cameras and LiDAR sensors on a robot that navigates around a school campus, which perfectly suits our desired setting.
Since 3D pose annotations are not provided in the original JRDB, we extracted sequences of 3D human pose from visible agents in omnidirectional images using the latest algorithm for 3D pose extraction from image~\cite{sun2022putting}.
We then ensure the reliability of 3D pose information by filtering and adjusting the extracted 3D poses based on 2D pose and 3D bounding box annotations.
% There is a human data set acquired by a robot called JRDB while moving around various environments. 
% This data provides detection and tracking results, camera images, and lidar sensor data.
% It provides 2D pose annotation, but does not provide the 3D pose required for human pose prediction.
% We have constructed our dataset by obtaining 3D human pose from the video sequence through the latest 3D pose estimation algorithm.
% To ensure its reliability, filtering \& adjusting were performed through comparison with the GT 2d annotation provided by JRDB.
As a result, we construct a real-world 3D human pose forecasting dataset, JRDB-GlobMultiPose (JRDB-GMP), where up to 24 agents exist for up to 5 seconds.
The proposed pose forecasting model is validated on both previous datasets and newly created JRDB-GMP dataset.
Our method shows state-of-the-art forecasting performance in both global and local accuracy metrics, not only on JRDB-GMP but also on all previous datasets.
Therefore, our contributions are as follows:

% Paragraph 5.
% - Summary 및 contribution 설명
%   - 1. Trajectory proposal 기반의 motion decoding 방식 제안, decoupling of global & local human motion
%   - 2. Past 3d motion 및 trajectory information의 그래프 기반 interaction modeling 방식 제안
%   - 3. Suggests a high-fidelity(?) dataset for multi-agent (6+), long-term (~7s) human motion forecasting
%   - 4. 제안된 방법론이 sota 달성

% 우리는 개발한 모델, 데이터셋을 여러 방면에서 검증하였다.
% 개발한 모델은 기존 pose prediction 데이터셋 뿐만 아니라 새롭게 만든 데이터셋에도 검증하였다.
% 또한 새롭게 만든 데이터 또한 기존의 prediction model 을 사용해 성능을 측정하였다.
% 실험 결과, 우리가 제안한 모델은 새롭게 파싱한 데이터에서 long-term / multi-agent 예측 성능을 SoTA 성능을 달성하였다.
% 흥미로운 점은 우리의 모델이 기존 데이터셋에도 sota 성능을 달성하였다.
% 따라서 우리의 contribution은 다음과 같다:
% long-term multi-agent human pose prediction 을 위한 global-trajectory conditioned prediction 방법을 개발함
% real-world 데이터로부터 신뢰성있는 3d human pose forecasting 데이터셋을 제안함
% 제안한 모델이 제안한 데이터셋 뿐만 아니라, 기존 데이터셋에서도 sota prediction 성능을 달성함
% 
% We have verified our trajectory-conditioned pose forecasting model and JRDB-GMP dataset in several ways.
% Our model is evaluated on both previous datasets (CMU-Mocap~\cite{cmumocap}, UMPM~\cite{6130396}, 3DPW~\cite{Marcard_2018_ECCV}) and our newly parsed dataset.
% Additionally, previous SOTA methods are also evaluated on the JRDB-GMP dataset to verify the authenticity of our new dataset.
% Our method targeting long-term and multi-agent pose forecasting achieves SOTA performance on the newly parsed JRDB-GMP dataset.
% Interestingly, our method also achieves SOTA performances on existing datasets.
% Therefore, our contributions are as follows:

- We propose a interaction-aware trajectory-conditioned pose forecasting method (\textit{T2P}) for long-term multi-agent 3D human pose forecasting.

- We propose a long-term, multi-agent real-world 3D human pose forecasting dataset which contains up to 24 persons and forecasts up to 5 seconds.

- We validate our \textit{T2P} model on both previous datasets and our new JRDB-GMP dataset. Our method achieves state-of-the-art forecasting performance on all datasets.
\section{Related works}
\label{sec:formatting}
\vspace{-3pt}
\begin{figure*}
    \includegraphics[width=\textwidth]{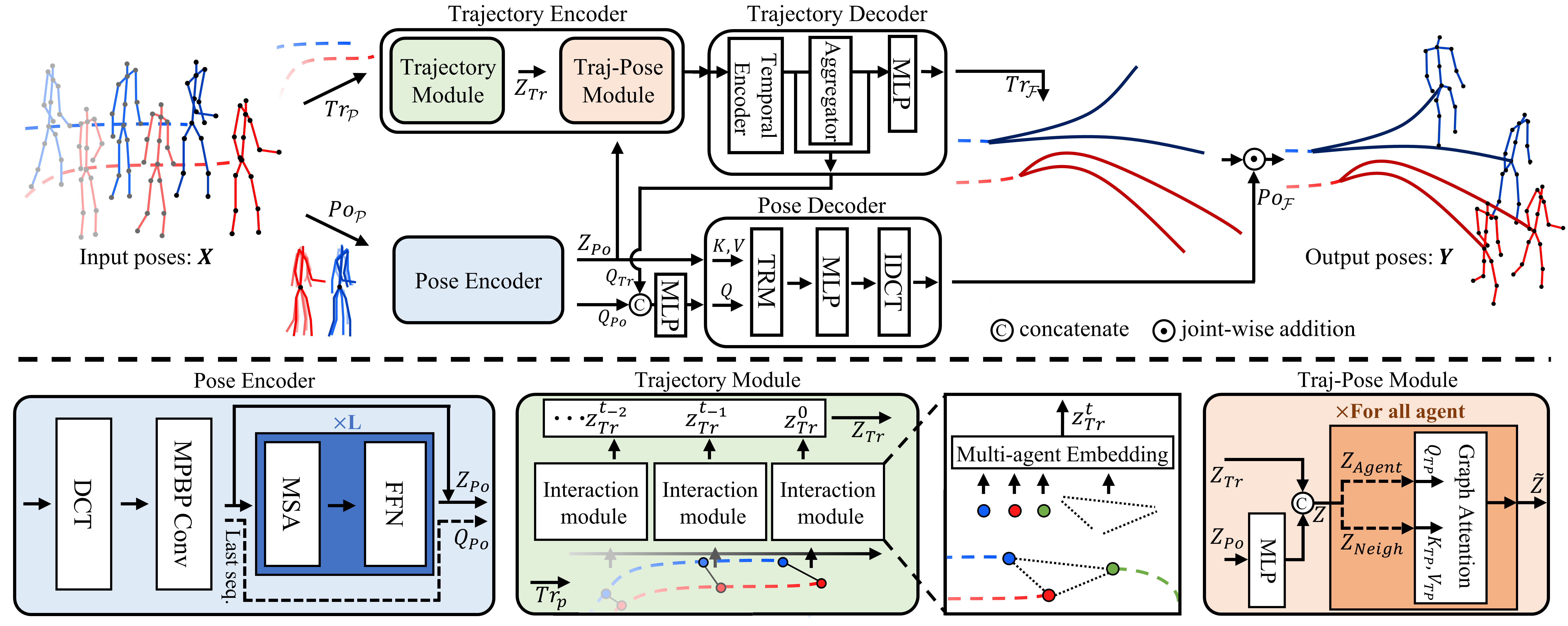}
    \caption{
    Illustration of our T2P framework. 
    We decompose global motion into global trajectory and local pose.
    Multi-modal global trajectory proposals are predicted from past global trajectory and local pose embeddings. 
    Then, future local poses are conditioned and forecasted on each trajectory proposal to compose the final human pose prediction.
    Predicted local poses are added to their mode-specific global trajectories in a joint-wise manner, obtaining the global human poses as the final output.
    }
    \label{method:fig2_framework}
    \vspace{-10pt}
\end{figure*}
\subsection{Human pose forecasting}
% Human Pose Forecasting 은 사람의 과거 pose의 history가 주어졌을 때 prediction horizon 동안의 미래 pose sequence 를 예측하는 것이다.
% Human pose는 사람 joint 들의 움직임으로 나타난다.
% 그 중 multi-agent 를 고려하거나, long-term 을 예측하는 것이 중요하다고 여겨져 왔다.
% ~~ 논문은 ~~ 방법을 사용했다.
% ~~ 논문은 ~~ 방법을 사용했다.
% 평가 데이터셋으로는 ~~가 주로 이용된다.
% 기존 데이터셋은 예측 길이가 3초 내로 짧은 horizon 을 갖는다.
% multi-agent 예측을 위해서는 cmu-mocap 이 사용되는데, 이 데이터셋은 사람이 3명까지 포함되어 있다.
% 따라서 기존의 multi-agent 논문에서는 개별의 scene 들을 임의로 합쳐서 3 명 이상의 사람이 존재하는 데이터셋을 만들었다.
% 하지만 이러한 방식은 agent 들간의 interaction이 제대로 반영되어있다고 할 수 없다.
Human Pose Forecasting involves predicting a future pose sequence with temporal length of a prediction horizon, given a historical pose sequence~\cite{gao2023decompose, mao2022contact, saadatnejad2023generic, ijcai2022p111, rahman2023best, parsaeifard2021learning, wang2021simple, Xing_2023_ICCV, Choudhury_2023_ICCV, Chen_2023_ICCV}.
In the early stage, methods were developed to forecast single person motion within a short timeframe ($\sim$ 1s)~\cite{chiu2019action, mao2019learning, sofianos2021space, Xu_2023_ICCV}.
However, to improve applicability on diverse downstream computer vision tasks, forecasts are to be made on multi-person poses~\cite{adeli2021tripod, adeli2020socially, Tanke_2023_ICCV} for longer prediction horizons~\cite{cao2020long, tanke2021intention}.
% The mainly used metrics are APE, JPE and FDE which quantify error to ground truth future.
% It is both common to predict a single future pose candidate or multiple futures~\cite{xu2023stochastic, Barquero_2023_ICCV} because predicting future inherently has stochastic nature. 
% As interaction complexity grows in these complex scenes, modeling the multi-modality of human motion becomes essential for a proficient forecast. 
% ill-pose된 problem이기 때문에 complexity가 높을수록 multi-modal 한 prediction을 이용하여 prediction을 잘 해야한다?
Forecasting future inherently involves a stochastic nature, and handling such multi-modality has been attempted by forecasting multiple future poses of a single agent~\cite{Barquero_2023_ICCV}.
However, comparatively marginal efforts have been employed in the more complex long-term multi-agent scenes~\cite{xu2023stochastic}.
Such absence is mostly due to the lack of a proper dataset.
The commonly used evaluation datasets are CMU-Mocap~\cite{cmumocap}, 3DPW~\cite{Marcard_2018_ECCV}, UMPM~\cite{6130396}, MuPoTS-3D~\cite{8490962}, all of which contain 2 agents at most in a given scene and  have short prediction horizons within 3 seconds.
% For multi-agent prediction, CMU-Mocap is often used, which includes up to three individuals.
Most recent research arbitrarily combines individual scenes to create datasets with more than three individuals~\cite{wang2021multi,peng2023trajectory,xu2023stochastic}.
However, such a synthetic approach does not account for authentic agent interactions.
% These limitations call for an attempt to handle the multi-modality of complex scenes in both model and dataset aspects.

\subsection{Trajectory prediction}
% Trajectory preiction은 object 의 과거 경로가 주어졌을 때, 그것의 미래 경로를 예측하는 것이다.
% 주로 자율 주행 환경에서 차량, 혹은 사람의 경로를 주로 예측하는 challenging 한 분야이다.
% Human pose forecasting과 다른 점은 차량이나 사람을 하나의 강체로 가정하고, 그것의 center point 의 움직임만을 예측한다는 것이다.
% Trajectory prediction 에서도 long-term 예측, 또는 multi-agent 예측이 중요하게 여겨져 왔다.
% 이 중 중요한 방향 중 하나는, Goal-conditioned prediction 방식이다.
% Goal-conditioned 방식은, 미래 trajectory 바로 예측하는 것이 아니라, prediction horizon 에서 최종 도착지점 (goal)을 먼저 예측한다.
% 이 때, 여러개의 goal 을 동시에 예측한 다음, 미래 경로를 각각의 goal 에 conditioned 하여 예측한다.
% 이렇게 하면 trajectory를 바로 예측하는 것보다 sample 의 mode 를 더 잘 나누어 학습할 수 있게 된다고 알려져 있다.
Trajectory prediction involves predicting the future path of an object given its past trajectory~\cite{zhu2023ipcc, xu2023uncovering, zhou2023query, rowe2023fjmp, choi2023r, aydemir2023adapt, jiang2023motiondiffuser, mao2023leapfrog, ngiam2022scene, Ye_2023_ICCV, park2023improving}.
% It is a challenging field primarily focused on predicting the paths of vehicles or individuals in autonomous driving environments.
Unlike human pose forecasting which aims to predict every joint position, trajectory prediction regards each agent as a point mass, typically the center of mass or center point of a detected bounding box.
Research in trajectory forecasting is interested in not only vehicles but also many types of agents including humans, cyclists, and so on~\cite{park2024t4p, li2022graph, xu2022remember, Zheng_2021_ICCV}.
% Trajectory forecasting is interested in not only vehicle, but also to many types of agents including human, cyclist, and so on~\cite{li2022graph, xu2022remember, Zheng_2021_ICCV}.
% typically forecasts the movement of their center point.
% Also in trajectory prediction, long-term prediction and multi-agent prediction have also been considered important~\cite{lee2022muse, }.
One substantial direction in research within this field is the Goal-conditioned prediction approach~\cite{Gu_2021_ICCV, Zhao_2021_ICCV, Mangalam_2021_ICCV}.
Goal-conditioned prediction approach first predicts the final destination within the prediction horizon with multiple goal proposals~\cite{lee2022muse, wang2023ganet}.
Then, a thorough future path is conditioned on each mode of the multi-modal proposals.
Compared to directly predicting full trajectories, the goal-conditioned approach follows a coarse-to-fine prediction and is effective in learning highly stochastic multi-modality of complex scenes~\cite{gu2021densetnt, zhou2022hivt, park2023leveraging}.
% better capturing sample modes

\subsection{Human pose estimation from image}
% Human pose estimation 은 이미지나 비디오로부터 사람의 pose 를 추정하는 것이다.
% Human pose estimation 역사 대략적으로 설명 (1)
% Human pose estimation 역사 대략적으로 설명 (2)
% Human pose estimation 역사 대략적으로 설명 (3)
% 우리는 jrdb 데이터 사용했는데, 이건 6(?) 개의 surround 2d camera 로 촬영한 것이다.
% ~~ 모델은 2D camera 이미지로부터 3D human pose 를 estimate 하는 모델이다
% 사람들의 크기에 따라 거리가 ambiguous 한 문제를 해결하였고, ~~한 특징이 있다.
% 카메라 파라미터를 가지고 3D human pose 를 global coordinate 로 보낼 수 있어서, jrdb 이미지에서 multi-person 이 interaction 하고 있는 pose의 global 좌표를 알아내기 위해 사용하였다.
% Human pose estimation involves estimating the pose of a person from images or videos.
% A brief overview of the history of human pose estimation (1).
% A brief overview of the history of human pose estimation (2).
% A brief overview of the history of human pose estimation (3).
% We utilized the JRDB dataset, which was captured by approximately six surround 2D cameras.
% The ~~ model is designed to estimate 3D human pose from 2D camera images.
% It addresses the problem of distance ambiguity based on the size of individuals and has ~~ feature.
% Leveraging camera parameters, it translates 3D human pose into global coordinates, enabling the determination of global coordinates for poses involving multi-person interactions in JRDB images.

Human pose estimation is the task of inferring the pose of a person from an image or a video~\cite{Zhang_2023_ICCV, You_2023_ICCV, Zhai_2023_ICCV, Jiang_2023_ICCV, Liu_2023_ICCV, Holmquist_2023_ICCV, Sun_2023_ICCV, Shi_2023_ICCV, Kan_2023_CVPR}. 
% In the early stage, hand-crafted feature and graphical models are used and to estimate 2D poses from single images.
Initial deep learning-based methods first utilized convolutional neural networks to estimate 2D and 3D poses from single or multiple images~\cite{Yu_2023_ICCV, Cin_2023_ICCV, Feng_2023_CVPR, Tang_2023_CVPR, Zhao_2023_CVPR}.
Recent approaches engage in more challenging tasks such as estimating 3D poses from monocular videos~\cite{Chai_2023_ICCV, Shan_2023_ICCV, Peng_2023_ICCV, Raychaudhuri_2023_ICCV, Shen_2023_CVPR, Qiu_2023_CVPR} using self-supervised learning and generative methods~\cite{Gong_2023_CVPR, Feng_2023_ICCV}.
Most recent methods estimate multi-person poses in a crowded environment with considerable occlusions~\cite{Zhou_2023_ICCV, Park_2023_ICCV}.
We account for the aforementioned need of a complex dataset by extracting 3D pose from images using these methods.
% To tackle the aforementioned dataset aspect for modeling multi-modal intentions in a complex environment, we use SOTA methods to construct a long-term multi-agent 3D pose dataset.
% In doing so, we use the 
Specifically, we use a monocular 3D pose estimation method \textit{BEV}~\cite{sun2022putting} to construct a 3D human motion forecasting dataset with long-term multi-agent characteristics from real-world image sequences.
\textit{BEV} robustly estimates human pose in a scale-ambiguous and crowded environment, reliably extracting 3D poses from the omnidirectional image sequences of JRDB dataset~\cite{vendrow2023jrdb}.

% To esimate exact 3D human pose in crowd real-world JRDB dataset~\cite{vendrow2023jrdb} which consists of five omnidirectional 2D cameras in both indoor and outdoor environments, we use pose estimation method.
% To extract reliable 3D poses from these omnidirectional image sequences, we use a SOTA method for monocular 3D pose estimation method \textit{BEV}~\cite{sun2022putting} that robustly estimates human pose in a scale ambiguous and crowded environment.

% The dataset contains multiple people interacting with each other and with objects. 
% Our model is a 3D human pose estimation model that takes 2D camera images as input and outputs 3D human poses. 
% Our model has two novel features: (a) it solves the ambiguity of depth estimation caused by different human sizes, and (b) it leverages temporal information to improve the accuracy and smoothness of pose estimation. 
% We use the camera parameters to transform the 3D human poses from camera coordinates to global coordinates. This allows us to obtain the global positions of multiple people who are interacting in the JRDB images.
\section{Method}
\subsection{Problem definition}
Multi-agent human pose forecasting aims to learn a mapping function between the observed 3D pose of $N_A$ agents composed of $J$ joints, $\textbf{\textup{X}} : \left \{ \textup{\textbf{x}}_{t}^{n, j} \right \}^{N_A, J}_{-T_p:0}$, and future pose $\textbf{\textup{Y}} : F\times\left \{ \textup{\textbf{x}}_{t}^{n, j} \right \}^{N_A, J}_{0:T_f}$ in global coordinates where $F$ denotes the number of modes.
Here, $T_p$ and $T_f$ are history length and prediction horizon while $\textup{\textbf{x}}_{t}^{n, j} = ( x^{n,j}_t, y^{n,j}_t, z^{n,j}_t )$ is 3D global coordinate of joint $j$ of agent $n$ at time $t$.
While the global position of joint is represented in $\textup{\textbf{x}}$, we additionally define local position $\textup{\textbf{p}}$.
The local position is defined in local coordinate of each agent, calculated by subtracting the global position of the hip joint of each agent.
%$\textup{\textbf{x}}^{n, \textup{hip}}$.
Therefore, local position of joint is defined as $\textbf{\textup{p}}^{n,j} = \textbf{\textup{x}}^{n,j} - \textbf{\textup{x}}^{n,\textup{hip}}$.
We define the trajectory of global hip joint position as global trajectory, $Tr : \left\{ \textbf{\textup{x}}^{n,\textup{hip}} \right\}^{N_A}$.
We also define local pose as local position of all joints, $Po : \left\{ \textbf{\textup{p}}^{n,\textup{j}} \right\}^{N_A, J}$.
% For both $Tr$ and $Po$, relative position to the previous timestep is used as input.
We denote past and future timesteps of global trajectory and local motion as $Tr_{\mathcal{P}},Tr_{\mathcal{F}}\in Tr$ and $Po_{\mathcal{P}},Po_{\mathcal{F}}\in Po$, where $\mathcal{P}$ and $\mathcal{F}$ respectively denotes past and future.

\subsection{Overall framework}
% In this section, we introduce our T2P model which performs human motion forecasting in a two-stage manner, namely trajectory prediction in global coordinates followed by local motion generation from generated trajectory proposals. 
% To better handle its multi-modality, w
We disentangle the overall human motion into global trajectories and local poses, as depicted in top left of Fig.~\ref{method:fig2_framework}.
% each depicted by the global trajectory of the central human body part (hip joint) and relative coordinates of other joints in reference to the hip joint.
Following a coarse-to-fine strategy, multiple global trajectories are first forecasted to model the coarse modes of global intentions.
Based on these forecasts, local pose predictions are conditioned on each mode to jointly constitute a thorough motion.
% , followed by local pose forecasts conditioned on each proposal of the multi-modal trajectories.
In doing so, our model is widely divided into two portions: Trajectory predictor consists of trajectory encoder and decoder and pose predictor consists of pose encoder and decoder.
Both predictors engage in the reciprocal exchange of both trajectory and pose information, facilitating the inference of cues between global and local motion.
% The first portion is for global trajectory prediction, where its graph attention-based interaction serves as an encoder and propagates the trajectory and motion embeddings, followed by MLP-based decoder for subsequent 3D trajectory prediction of hip joints.
% % A MLP-based decoder subsequently predicts 3D trajectories of each agent's hip joint from the updated embeddings.
% As for the local pose encoder model, a transformer model encodes the intra-agent input joint relations in local hip coordinates. 
% Then, a transformer-based decoder conditions pose embedding on propagated multi-modal trajectory embeddings to forecast mode-specific future local motions.
% After predicting each local motion, joint-wise addition of local motion to global hip joint motion is performed to each mode to obtain the final global pose forecasts.
The detailed methods of each stage are described below:

\subsection{Model structure}
\subsubsection{Pose encoder}
Unlike the holistic approach of previous works that encode and decode all agents' joint motions in global coordinates, our pose encoder encodes the pose dynamics in local coordinates.
In addition, our pose encoder only considers intra-agent joint interaction.
As a result, the encoded pose embedding represents agent-specific local motion, containing insights on global intent.
We follow our baseline~\cite{peng2023trajectory} and construct the encoder with Multi-Person Body-Part (MPBP) module and transformer networks.
As depicted in Fig.~\ref{method:fig2_framework}, body part sequences are constructed in frequency domain, followed by intra-agent attention-based encoding of the body parts to acquire pose embedding $Z_{Po}$.

% In doing so, we first convert the local motion into frequency domain via discrete cosine transform 
% for efficient training.
% Then, we 
% and follow the approach of our baseline~\cite{peng2023trajectory}.
% and encode the intra-agent local pose in a body part-wise manner for simplicity and generalization.
% $N$ agents' past local motion $X\in\mathbb{R}^{N\times T\times J \times C}$ is first encoded into 5 body part features, where T, J, and C represent the number of timesteps, joints, and 3 spatial dimensions (x,y,z), respectively.
% Each body part is spatially and temporally encoded into a Multi-person Body-Part (MPBP) sequence via agent-wise convolutional layers in the temporal dimension.
% Then, we inject sinusoidal temporal and identity encoding into the MPBP sequence prior to layers of multi-head self-attention (MSA-PE) transformer. 
% The output of stacked transformer layers from MPBP sequence input is acquired as the pose embedding \textup{$Z_{Po}\in\mathbb{R}^{p\times d_{Po}}$}, and $p=N_{A}\times N_{body part}\times (T-k)$ where $k$ is convolution kernel size.
% Only intra-agent joint interaction is considered during both MPBP Conv and MSA-PE transformer operations.
% , namely $X_{MPBP}\in\mathbb{R}^{N\times L\times B \times D}$ where $L$ is the resulting temporal length ($L=(T-L_{kernel}+1)/stride$), $B$ the number of body parts, and $D$ the feature dimension. 

\subsubsection{Trajectory module}
Trajectory module aims to extract embeddings from the agents' past global trajectory.
%, acquired by graph-based interaction. 
Using an encoder structure from~\cite{zhou2022hivt}, multi-agent interaction-based trajectory embedding $Z_{Tr}$ is extracted which contains rudimentary insight on global intent.
Interaction between agent trajectories is represented based on the reference agent $i$'s global trajectory segment vector $\textbf{\textup{v}}_t^i = \textbf{\textup{x}}_{t}^{i,\textup{hip}} - \textbf{\textup{x}}_{t-1}^{i,\textup{hip}}$.
For rotational invariance, neighbor actor $j$'s vector is normalized by the reference vector's orientation at latest timestep t=0.
Separate MLP layers then compute the reference agent and neighboring agent embeddings $z_{Tr_i}^t, z_{Tr_j}^t$ as follows:

\begin{equation}
\begin{aligned}
\textup{$z_{Tr_i}^t = \phi_{ref}(R^T_{\textit{i}} \textbf{\textup{v}}_t^i )$} \\
\textup{$z_{Tr_j}^t = \phi_{nbr}([R^T_{\textit{i}}( \textbf{\textup{v}}_t^j ), R^T_{\textit{i}}( \textbf{\textup{v}}_t^i )])$} \\
\end{aligned}
\end{equation}

\noindent where $\phi_{ref}$ and $\phi_{nbr}$ are different MLP blocks, $R_i \in \mathbb{R}^{3 \times 3}$ is the rotation matrix of agent $j$ against agent $i$, $\left [ \cdot , \cdot \right ]$ is concatenation.
The resulting agent-specific reference and neighbor embeddings constitute trajectory embedding $z_{Tr}^t$.

\subsubsection{Traj-pose module}
Human maneuver contains various dynamic activities characterized by the agent's multi-modal intents.
Auxiliary human motion such as arm gesture, rotational orientation of upper body and head implies the agent's intent in global motion.
In that sense, harvesting meaningful insights from past local joint motion helps proficient modeling of coarse multi-modality as future trajectory proposals.
Therefore, we propose Traj-Pose Module that fuses agent-wise embeddings of both trajectory and pose to fully utilize these information in modeling global intentions.

First, MLP is used to match the temporal domain of pose embedding $Z_{Po}$ to that of $Z_{Tr}$, after which both are concatenated as agent-wise traj-pose embedding $Z$.
\begin{equation}
\begin{aligned}
\textup{$Z = [Z_{Tr}, \phi_{MLP}(Z_{Po})]$} \\
\end{aligned}
\end{equation}
The resulting $Z$ is comprised of agent and timestep-respective trajectory and pose embeddings: $z_{i}^t,z_{j}^t \in z^t \in Z$
Then, $\widetilde{Z}$ is acquired from the graph attention with an agent-wise update where each agent embedding $z_{i}^t$ and its neighbor embedding $z_{j}^t$ are used as query and key/value.
% \begin{equation}
% \begin{aligned}
% \textup{$q_i^t=W^Q z_{i}^t,  k_j^t=W^K z_{j}^t,  v_j^t=W^V z_{j}^t$}\\
% \end{aligned}
% \end{equation}
% Where $W^Q$, $W^K$, $W^V$ are learnable matrices for respective linear projection of embedding into tranformed dimension $d_k$. 
Similar to trajectory interaction encoder of HiVT~\cite{zhou2022hivt}, the graph attention operation is operated as follows:
\begin{equation}
\begin{aligned}
\textup{$\alpha_i^t = \textrm{softmax}(\frac{q_i^{t\top}}{\sqrt{d_k}}\cdot[\{k^t_j \}_{j\in N_i}])$,}\\
\textup{$m^t_i = \sum_{j\in N_i} \alpha_i^t v_j^t$,}\\
\textup{$g_i^t = \textrm{sigmoid}(W^{\textrm{gate}}[z_i^t,m_i^t])$,}\\
\textup{$\widetilde{z}_i^t = g_i^t \odot W^{\textrm{self}}z_i^t + (1-g_i^t) \odot m_i^t$}\\
\end{aligned}
\end{equation}
\noindent where $N_i$ is a set of agent \textit{i}'s neighbors, $W^{gate}$ and $W^{self}$ are learnable matrices, and $\odot$ is element-wise product. 
% Since such graph operation is operated by each timestep, a temporal encoder is used as a temporal encoder to integrate the embedding in the temporal dimension.
% A multi-head self-attention temporal encoder is used as the temporal encoder which utilizes an extra learnable token and temporal masks similar to BERT~\cite{devlin2018bert}.

% While Trajectory Module and Traj-Pose Module are separately illustrated in Fig.~\ref{method:fig2_framework} to elucidate our framework, they operate simultaneously prior to the temporal encoder.

\subsubsection{Trajectory decoder}
% To sufficiently consider the interaction between agents using their trajectory and pose information, the trajectory/pose embedding $\widetilde{Z}$ acquired from Traj-Pose module goes through an additional layer of graph operation as an aggregator. 
Trajectory is forecasted from the output of trajectory encoder which encodes both past global trajectory and local pose information.
Since its graph operation is operated by each timestep, a temporal encoder is used as a temporal encoder to integrate $\widetilde{Z}$ in the temporal dimension.
A multi-head self-attention temporal encoder is used as the temporal encoder.
%which utilizes an extra learnable token and temporal masks similar to BERT~\cite{devlin2018bert}.
Aggregator then takes into account variations in local coordinate frames to accurately represent geometric relationships within the global coordinate system via a graph operation.
MLP is subsequently applied to span embedding $F$ times for multi-modal prediction, which is residually added to the $\times F$ repeated embedding before the aggregator.
Finally, another MLP is used to extract multi-modal future global trajectory proposals of hip joint $Tr_{\mathcal{F}}\in\mathbb{R}^{F\times T_f \times3}$.
The multi-modal embedding is also passed onto the Pose Decoder to forecast local poses.

\subsubsection{Pose decoder}
Future human pose depends on past human poses and global intention. 
The pose decoder is designed to consider these factors while generating local poses via mode-specific trajectory conditioning.
% Such mode-specific forecasts consolidate the modeling of multi-modality.
% Overall, global trajectories and their respective local poses constitute the multi-modal human motion.
A transformer (TRM) decoder is used to decode local motions, where pose embedding $Z_{Po}$ is used as key/value and concatenation of trajectory and pose query.
%as final query after reducing its dimension into $d_{Po}$ via MLP: $\phi_{MLP}\in\mathbb{R}^{d_{Po}\times(d_{Po}+d_{Tr})}$.
%
\begin{equation}
\begin{aligned}
\textup{$Q =  \phi_{MLP}([Q_{Po}, Q_{Tr}])$}, K = Z_{Po}, V = Z_{Po}\\
\end{aligned}
\end{equation}
Using both pose and trajectory queries, past pose embedding $Z_{Po}$ is conditioned on both MPBP sequence at t=0 and the multi-modal trajectory proposals which contain global intent.
% Using both pose and trajectory queries, past pose embedding $Z_{Po}$ is conditioned on the agent's global intent inferred by both MPBP sequence at t=0 and the multi-modal trajectory proposals.
% The multi-modal local pose forecast in frequency domain is acquired from multi-head attention-based transformer operation of $Q\in\mathbb{R}^{F \times N_A\times d_{Po}}, K,V\in\mathbb{R}^{p\times d_{Po}}$ followed by MLP.
Subsequently, inverse discrete cosine transform (idct) is applied to convert the future pose proposals from frequency domain to local coordinate domain, $Po_{\mathcal{F}}$.
The final multi-modal future pose in global coordinates is acquired as Eq.~\ref{eq:joint_addition} where $\oplus$ is a joint-wise addition operation.
\begin{equation}
\begin{aligned}
\textup{$\textbf{Y}=Tr_{\mathcal{F}} \oplus Po_{\mathcal{F}}, \quad \textbf{Y}\in \mathbb{R}^{F\times N_A \times T_f \times 3}$}\\
\end{aligned}
\label{eq:joint_addition}
\end{equation}
%
% Where $\oplus$ denotes a joint-wise addition operation.
% Overall, our approach better handles multi-modality by initially forecasting multiple modes of coarse global trajectory, followed by mode-specific conditioning and local motion forecasting.

\subsection{Training objective}
Both objectives of global trajectory and local pose forecasting are trained jointly. 
For both global trajectory and local pose prediction, $L2$ loss is propagated to the mode with minimal $L2$ distance with the ground truth. 
\begin{equation}
\begin{aligned}
\textup{$L_{Tr} = \sum_{n=1}^{N_A}\sum_{t=1}^{T_f} \lVert \widetilde{y}_{Tr,n}^t - \hat{y}_{Tr,n}^t \rVert$}\\
\textup{$L_{Po} = \sum_{n=1}^{N_A}\sum_{t=1}^{T_f}\sum_{j=1}^{J-1} \lVert (\widetilde{y}_{Po,n}^{t,j} - \hat{y}_{Po,n}^{t,j}) \rVert$}\\
L = L_{Tr} + L_{Po}
\end{aligned}
\end{equation}

% The MPBP sequence is also used as a query for the pose decoder network. 
% The pose decoder network is designed to predict future poses conditioned on the agents' past trajectory and their pose at $t=0$, for we believe these two are the most crucial factors that determine their future behavior.
% In doing so, the last temporal sequence of encoded MPBP sequence is passed onto the pose decoder as pose query. 
% \begin{figure}[H]
% \includegraphics[width=0.5\textwidth]{figs/figure_3_v3.PNG}
% \vspace{-20pt}
% \caption{Example scenes from the JRDB-GMP dataset, illustrating its long-term, multi-agent nature.}
% \label{experiment:fig3_datasetExamples_motion}
% \end{figure}

\definecolor{Gray}{rgb}{0.9,0.9,0.9}

\begin{table}
\centering
\caption{Comparison of statistics between existing human pose forecasting datasets and newly proposed JRDB-GMP dataset.}
\vspace{-5pt}
\label{tab:dataset_comparison}
\resizebox{\linewidth}{!}{
% \begin{tabular}{c|ccccc}
% \toprule
%                  & \multicolumn{4}{c}{Dataset}                       \\ \cline{2-5} 
%                  & \makecell{CMU \\ -Mocap \\ (UMPM)} & \makecell{MuPots \\-3D} & 3DPW & \makecell{JRDB\\ -GlobMultiPose\\1s/2s\;\;\;\;\;   2s/5s} \\ \hline
%                  % & & & & \makecell{1s/2s} \makecell{2s/5s} \\ \hline
% Duration (s)     & 4000      & 267       & 1700    & 1863 &                   \\ \hline
% location \#      & -         & 20        & -       & 27   &                   \\ \hline
% sample \#        & 13000     & 192       & 432     & 1153 & 4593              \\ \hline
% \makecell{avg. \\ agent \#}    &3 &3 &2 &6.8 &6.8                             \\ \hline
% \makecell{med. \\ agent \#}    &3 &3 &2 &5   &5                               \\ \hline
% \makecell{max \\ agent \#}     &3 &3 &2 &24  &22                              \\ \hline
% \makecell{avg. \\ speed (m/s)} &0.3 &0.26 &0.57 &0.46 &0.38                   \\ \hline
% \makecell{avg. \\ disp. (m)}   &0.63 over 3s &0.55 over 3s &1.13 over 2.4s &0.64 over 3s &0.79 over 7s                    \\ \hline
% \makecell{max \\ disp. (m)}   &4.62 over 3s &2.45 over 3s &      &                    \\ \bottomrule
% \end{tabular}
\begin{tabular}{cccccc}
\hline
\multirow{3}{*}{}                    & \multicolumn{5}{c}{Dataset}                                                                                                                                                                                                                                                             \\ \cline{2-6} 
                                     & \multicolumn{1}{|c|}{\multirow{2}{*}{\begin{tabular}[|c|]{@{}c@{}}CMU-Mocap \\ (UMPM)\end{tabular}}} & \multicolumn{1}{c|}{\multirow{2}{*}{\begin{tabular}[c]{@{}c@{}}MuPoTs\\ -3D\end{tabular}}} & \multicolumn{1}{c|}{\multirow{2}{*}{3DPW}} & \multicolumn{2}{c}{\cellcolor{Gray} JRDB-GMP} \\ \cline{5-6} 
                                     & \multicolumn{1}{c|}{}                                                                                & \multicolumn{1}{c|}{}                                                                      & \multicolumn{1}{c|}{}                      & \multicolumn{1}{c|}{\cellcolor{Gray} 1s/2s}   & \cellcolor{Gray} 2s/5s   \\ \hline
\multicolumn{1}{c|}{Duration (s)}    & \multicolumn{1}{c|}{4000}                                                                            & \multicolumn{1}{c|}{267}                                                                   & \multicolumn{1}{c|}{1700}                  & \multicolumn{2}{c}{\cellcolor{Gray} 1863}               \\ \hline
\multicolumn{1}{c|}{Location \#}      & \multicolumn{1}{c|}{-}                                                                               & \multicolumn{1}{c|}{20}                                                                    & \multicolumn{1}{c|}{-}                     & \multicolumn{2}{c}{\cellcolor{Gray} 27}                 \\ \hline
\multicolumn{1}{c|}{Sample \#}        & \multicolumn{1}{c|}{13000}                                                                           & \multicolumn{1}{c|}{192}                                                                   & \multicolumn{1}{c|}{432}                   & \multicolumn{1}{c|}{\cellcolor{Gray} 1153}    & \cellcolor{Gray} 4593    \\ \hline
\multicolumn{1}{c|}{avg. agent }    & \multicolumn{1}{c|}{3}                                                                               & \multicolumn{1}{c|}{3}                                                                     & \multicolumn{1}{c|}{2}                     & \multicolumn{1}{c|}{\cellcolor{Gray} 6.8}     & \cellcolor{Gray} 6.8     \\ \hline
\multicolumn{1}{c|}{med. agent \#}    & \multicolumn{1}{c|}{3}                                                                               & \multicolumn{1}{c|}{3}                                                                     & \multicolumn{1}{c|}{2}                     & \multicolumn{1}{c|}{\cellcolor{Gray} 5}       & \cellcolor{Gray} 5       \\ \hline
\multicolumn{1}{c|}{max agent \#}     & \multicolumn{1}{c|}{3}                                                                               & \multicolumn{1}{c|}{3}                                                                     & \multicolumn{1}{c|}{2}                     & \multicolumn{1}{c|}{\cellcolor{Gray} 24}      & \cellcolor{Gray} 22      \\ \hline
\multicolumn{1}{c|}{avg. vel. (m/s)} & \multicolumn{1}{c|}{0.3}                                                                             & \multicolumn{1}{c|}{0.26}                                                                  & \multicolumn{1}{c|}{0.57}                  & \multicolumn{1}{c|}{\cellcolor{Gray} 0.46}    & \cellcolor{Gray} 0.38    \\ \hline
\multicolumn{1}{c|}{avg. disp.(m)}   & \multicolumn{1}{c|}{0.63}                                                                            & \multicolumn{1}{c|}{0.55}                                                                  & \multicolumn{1}{c|}{1.13}                  & \multicolumn{1}{c|}{\cellcolor{Gray} 0.64}    & \cellcolor{Gray} 0.79    \\ \hline
\multicolumn{1}{c|}{max. disp.(m)}   & \multicolumn{1}{c|}{4.62}                                                                            & \multicolumn{1}{c|}{2.45}                                                                  & \multicolumn{1}{c|}{10.71}                  & \multicolumn{1}{c|}{\cellcolor{Gray} 8.44}    & \cellcolor{Gray} 11.0    \\ \hline
\end{tabular}%
}
\vspace{-4pt}
\end{table}

\begin{figure}[t]
\includegraphics[width=0.5\textwidth]{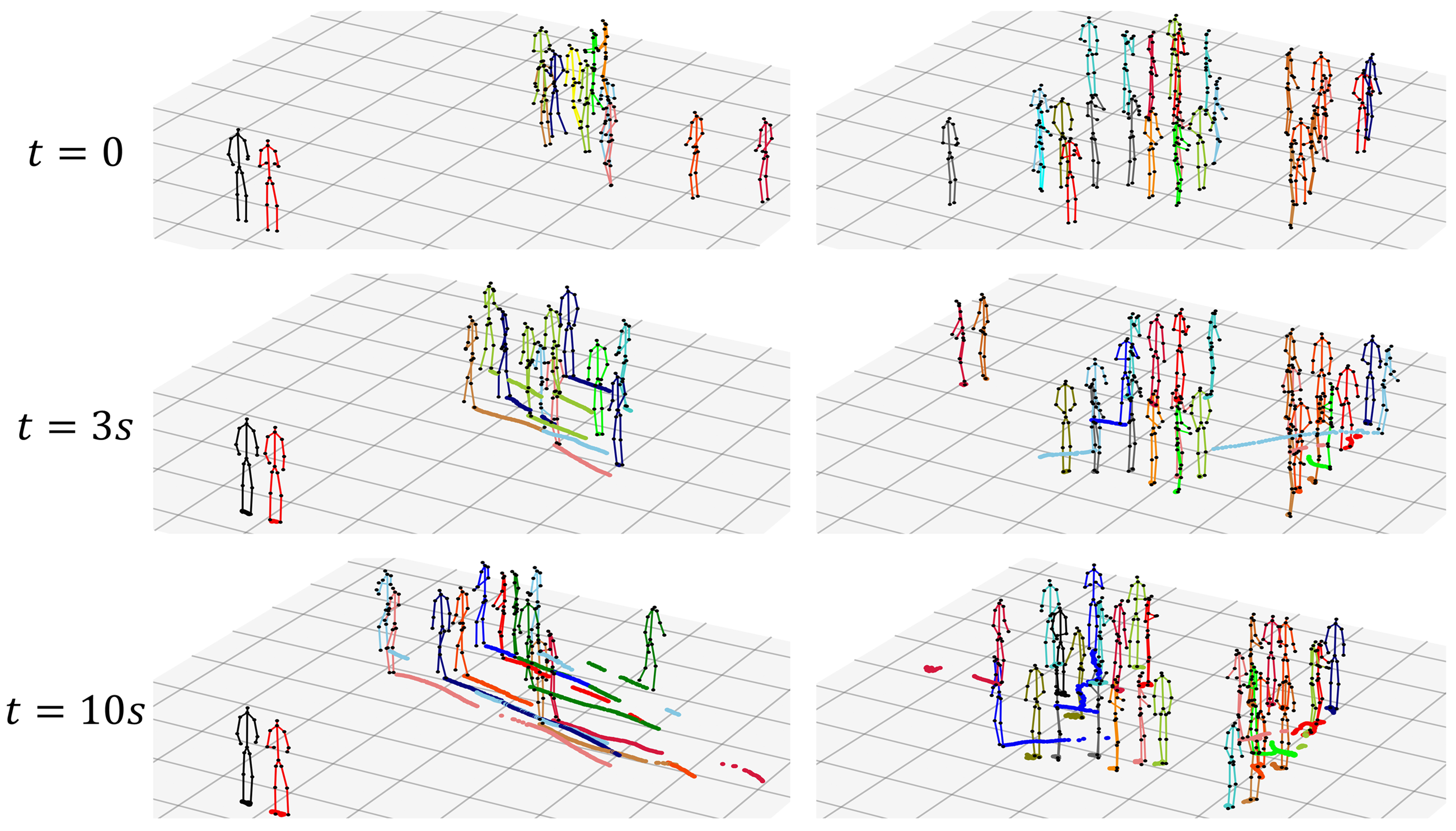}
\vspace{-20pt}
\caption{Example scenes from the JRDB-GMP dataset, illustrating its long-term, multi-agent nature.}
\label{experiment:fig3_datasetExamples_motion}
\vspace{-9pt}
\end{figure}
\begin{figure}[t]
\vspace{-3pt}
\includegraphics[width=0.5\textwidth]{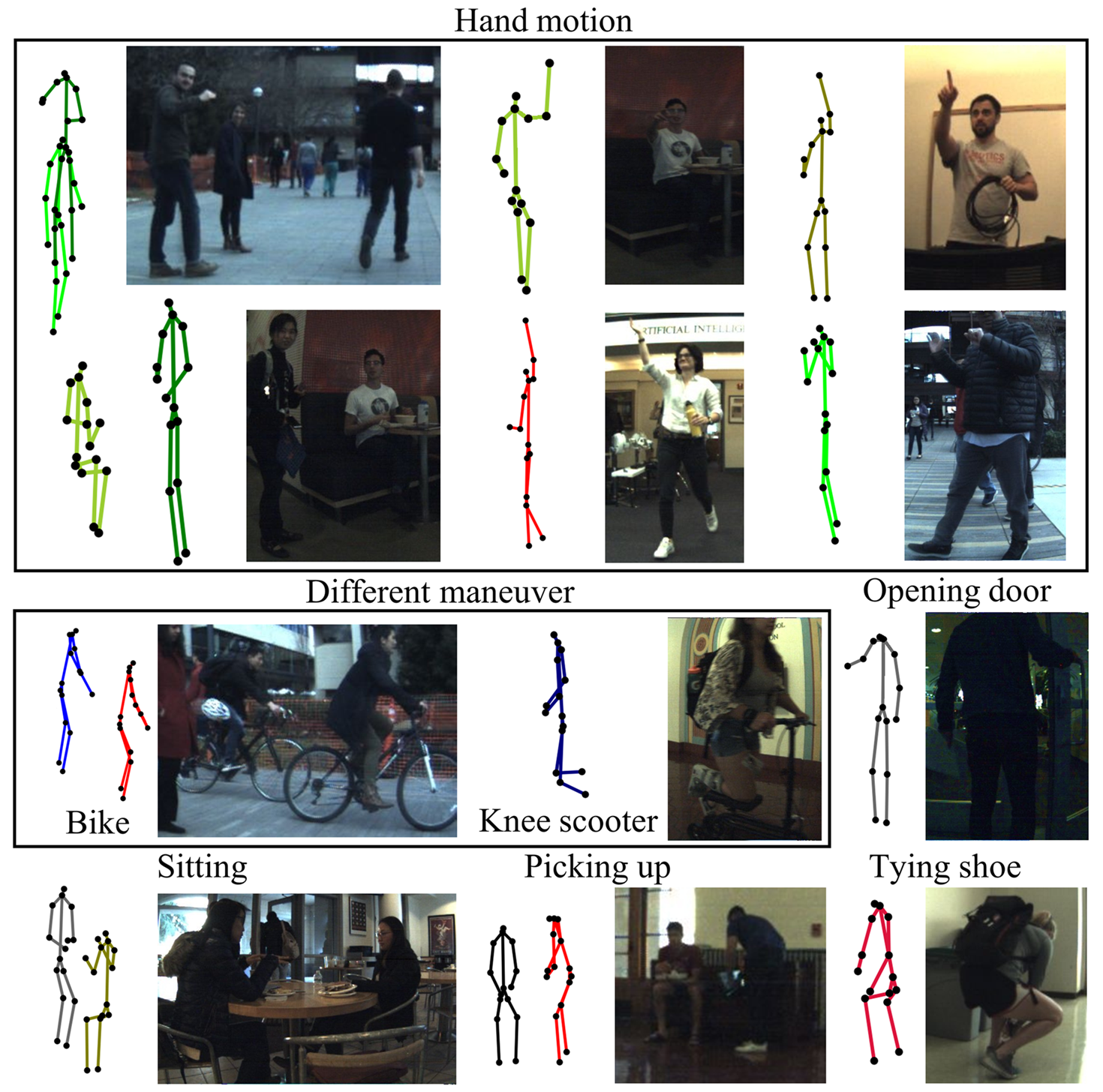}
\caption{Various motions from the JRDB-MultiGlobPose dataset, providing rich motion queues for inter-agent interaction inference.}
\vspace{-10pt}
\label{experiment:fig3_datasetExamles_pose}
\vspace{-5pt}
\end{figure}

\begin{table*}[t]
\caption{Quantitative comparison of our method to previous methods on CMU-mocap (UMPM), 3DPW, and JRDB-GlobMultiPose datasets with number of prediction modes ($F$) as 6. Lower is better for all metrics. The best results are marked in \textbf{bold}.}
\vspace{-10pt}
\label{tab:main_results}
\scriptsize
\centering
\resizebox{0.82\textwidth}{!}{
\begin{tabular}{ll|cc|cc|cccc}
\toprule
\multicolumn{2}{c|}{Dataset}             & \multicolumn{2}{c|}{\begin{tabular}[c]{@{}c@{}}CMU-mocap\\ (UMPM)\end{tabular}} & \multicolumn{2}{c|}{3DPW}       & \multicolumn{4}{c}{\begin{tabular}[c]{@{}c@{}}JRDB-\\ GlobMultiPose\end{tabular}} \\ 
\hline
\multicolumn{2}{c|}{In/out length (s)}   & \multicolumn{2}{c|}{1/2}                                                        & \multicolumn{2}{c|}{0.8/1.6}    & \multicolumn{2}{c|}{1/2}                               & \multicolumn{2}{c}{2/5}        \\ 
\hline
\multicolumn{2}{c|}{Evaluation time (s)} & 1                                      & 2                                      & 0.8            & 1.6            & 1               & \multicolumn{1}{c|}{2}               & 2.5               & 5              \\ 
\hline
\multirow{4}{*}{JPE}     & MRT~\cite{wang2021multi}           & 164.7                            & 280.1                            & 159.1          & 251.2    & 259.3           & \multicolumn{1}{c|}{349.3}           & 438.4           & 474.0    \\
& JRT~\cite{xu2023joint}           & 168.5                                  & 316.9                                  & 181.9          & 287.3          & 237.9     & \multicolumn{1}{c|}{373.1}           & 351.9     & 538.8          \\
& TBIFormer~\cite{peng2023trajectory}     & 170.0                                  & 290.9                                  & 153.9    & 265.8          & 257.1           & \multicolumn{1}{c|}{339.3}     & 443.2           & 481.3          \\
\rowcolor{tablegray}
& Ours          & \textbf{152.4}                         & \textbf{262.7}                         & \textbf{142.6} & \textbf{236.2} & \textbf{224.0}  & \multicolumn{1}{c|}{\textbf{301.4}}  & \textbf{341.6}  & \textbf{390.4} \\ 
\hline
\multirow{4}{*}{APE}     & MRT~\cite{wang2021multi}           & 127.0                                  & 164.4                                  & 117.9          & 153.2          & 72.3            & \multicolumn{1}{c|}{87.3}            & 88.5      & 101.9    \\
& JRT~\cite{xu2023joint}           & 121.2                            & 181.6                                  & 133.4          & 178.0            & 112.6           & \multicolumn{1}{c|}{154.3}           & 96.7            & 120.2          \\
& TBIFormer~\cite{peng2023trajectory}     & 125.1                                  & 160.8                            & 115.4    & 152.7    & \textbf{70.6}   & \multicolumn{1}{c|}{\textbf{83.3}}   & 88.2   & 102.9          \\
\rowcolor{tablegray}
& Ours          & \textbf{114.4}                         & \textbf{151.7}                         & \textbf{114.6} & \textbf{150.0} & 70.8      & \multicolumn{1}{c|}{\textbf{83.3}}      & \textbf{82.2}   & \textbf{94.7}  \\ 
\hline
\multirow{4}{*}{FDE}     & MRT~\cite{wang2021multi}           & 99.6                             & 204.7                            & 102.7    & 185.3    & 235.2           & \multicolumn{1}{c|}{325.2}           & 418.2           & 454.8    \\
& JRT~\cite{xu2023joint}           & 117.7                                  & 250.8                                  & 133.7          & 235.4          & 211.4     & \multicolumn{1}{c|}{337.4}           & 318.5     & 497.2          \\
& TBIFormer~\cite{peng2023trajectory}     & 112.1                                  & 228.5                                  & 106.7          & 215.9          & 232.4           & \multicolumn{1}{c|}{314.6}     & 423.9           & 458.8          \\
\rowcolor{tablegray}
& Ours          & \textbf{88.7}                          & \textbf{188.9}                         & \textbf{74.1}  & \textbf{158.2} & \textbf{194.7}  & \multicolumn{1}{c|}{\textbf{271.5}}  & \textbf{313.9}  & \textbf{361.0} \\ \bottomrule
\end{tabular}
}
\vspace{-13pt}
\end{table*}

\subsection{JRDB-GMP dataset}
% As explained in the introduction section, we compose a unique dataset for long-term (3s+) multi-agent (6+) 3D human pose forecasting in a real-world environment from JRDB\cite{vendrow2023jrdb}.
Due to the absence of existing long-term (3s+) multi-agent (6+) dataset, we compose a unique 3D human pose forecasting dataset in a real-world environment from JRDB~\cite{vendrow2023jrdb}. 
The original JRDB dataset is constructed by a moving robot that records human activity around a school campus using 5 omnidirectional cameras and LiDAR.
Image sequences along with 2D pose annotation and 3D bounding box annotations are provided in the original dataset.
% of this dataset are used to extract and refine the sequences of 3D human poses.
However, since 3D human pose annotations are unavailable, we separately parse accurate 3D human pose from provided inputs and annotations.
First, a SOTA monocular 3D pose extraction method \cite{sun2022putting} is used to extract raw 3D joint positions from image sequences.
Then, 2D pose and 3D bounding box annotations are used to refine the raw joint positions and minimize noise.
We use 2D pose annotations to initially filter out the 3D poses with noise. 
With camera parameters and refined 3D pose, we project it on 2D image plane, then L2 distance between projected 2D pose and GT 2D pose annotation is calculated.
If the mean L2 distance per each agent at a time stamp is over a threshold, that instance is filtered out.
2D pose annotations are also projected in 3D space to refine the remaining 3D poses, ensuring the accuracy of the 3D pose information of our JRDB-GMP dataset.
% Finally, 3D bounding boxes are used to adjust the central position of 3D pose. 
% We also use odometry information to offset the robot's ego-motion.
Further details are elaborated in the supplementary materials.

Figure~\ref{experiment:fig3_datasetExamples_motion} visualizes some scenes of the constructed dataset.
Accurate extraction of 3D poses has been made even with considerable occlusion via the use of 2D poses.
% Long-term multi-agent interaction is also well exhibited.
The dataset includes agents with both long and short traverse distances and rich inter-agent interactions in both trajectory and local pose aspects.
Figure~\ref{experiment:fig3_datasetExamles_pose} illustrates diverse local poses included in the dataset, which serve as motion cues of inter-agent interaction.
% Both two figures confirm that our method extracts accurate 3D multi-human pose even in a crowded environment.
Both figures confirm our method's accuracy in extracting 3D multi-human pose, even in crowded environments.
Table~\ref{tab:dataset_comparison} scrutinizes the statistics compared to previously used datasets.
Compared to earlier datasets, the average number of agents is more than twice as high. 
In addition, comparing JRDB-GMP 1s/2s to CMU-Mocap and MuPoTs datasets, JRDB contains more diverse and longer motion as shown by a similar magnitude of average displacement but longer maximum displacement.
% As for the 3DPW dataset, its large maximum displacement is due to the drift of agents caused by camera motion.

\section{Experiment}
\subsection{Dataset}
We test our model on three datasets: CMU-Mocap (UMPM)~\cite{cmumocap,6130396},
%parsed by our baseline~\cite{peng2023trajectory}, 
3DPW~\cite{Marcard_2018_ECCV}, and our JRDB-GMP. 
Although our model is designed to forecast human poses in a long-term multi-agent environment, we also report experimental results on previous benchmark datasets with simpler scenes.
Mocap-UMPM is a mixed dataset of Mocap and UMPM containing synthesized human interaction between three agents~\cite{peng2023trajectory}. 
% Both Mocap and UMPM recorded human motion using IMU sensors in a lab environment.
3DPW is a dataset with 2 agents traversing a real-world environment.
%with motion capture with IMU sensors. 
We report the test results on each after separate training on respective datasets.

\subsection{Metrics}
We use the following widely-used metrics. 
For a detailed definition, please refer to the supplementary material. \\
\textbf{APE}: Aligned mean per joint Position Error is used as a metric to evaluate the forecasted local motion.
$L2$ distance of each joint in the hip joint coordinate is averaged over all joints for a given timestep.\\
\textbf{FDE}: Final Distance Error evaluates the forecasted global trajectory by calculating the $L2$ distance of a given timestep. \\
\textbf{JPE}: Joint Precision Error evaluates both global and local predictions by mean $L2$ distance of all joints for a timestep.

\subsection{Implementation details}
We train our model on a single A6000 GPU. 
2 layers of pose encoder transformer are stacked, followed by 2 layers of transformer in pose decoder.
Embedding dimensions of 96 and 128 are used for trajectory and pose embeddings, respectively.
The transformed key, value dimension of 64 is used for all transformer architectures.
A learning rate of 0.003 is used with an AdamW optimizer with weight decay.
Further details can be found in the supplementary materials.

\begin{table}[ht!]
\centering
\footnotesize
\caption{Short-term prediction results on CMU-Mocap (UMPM) dataset, where 1s of poses are forecasted given 2s of poses.}
\vspace{-10pt}
\label{tab:ablation_short_term}
\resizebox{\linewidth}{!}{
\begin{tabular}{l|p{7.5pt}p{6pt}p{9pt}|p{7.5pt}p{7.5pt}p{9pt}|p{8pt}p{7.5pt}p{9pt}}
\toprule
\multicolumn{1}{c|}{Metric}        & \multicolumn{3}{c|}{JPE} & \multicolumn{3}{c|}{APE} & \multicolumn{3}{c}{FDE} \\ \hline
\multicolumn{1}{c|}{Time (s)} & 0.2    & 0.6    & 1.0    & 0.2    & 0.6    & 1.0    & 0.2    & 0.6   & 1.0    \\ \hline
MRT                      & 64.5   & 152  & 217  & 49.8   & 110   & 140  & 39.4   & 97.9  & 153  \\
JRT                      & \textbf{31.5}   & 104  & 173  & \textbf{28.7}   & 85.9   & 125  & 17.7   & 63.9  & 120  \\
\footnotesize{TBIFormer}                          & 37.4   & 104  & \textbf{158}  & 32.8   & 85.8   & 119  & 23.3   & 63.7  & 104  \\
 \rowcolor[rgb]{0.9,0.9,0.9} Ours                               & 37.8   & \textbf{102}  & \textbf{158}  & 33.8   & \textbf{84.4}   & \textbf{116}  & \textbf{14.9}   & \textbf{49.1}  & \textbf{92.6}   \\ \bottomrule
\end{tabular}
}
\vspace{-14pt}
\end{table}

\begin{figure*}
\includegraphics[width=\linewidth]{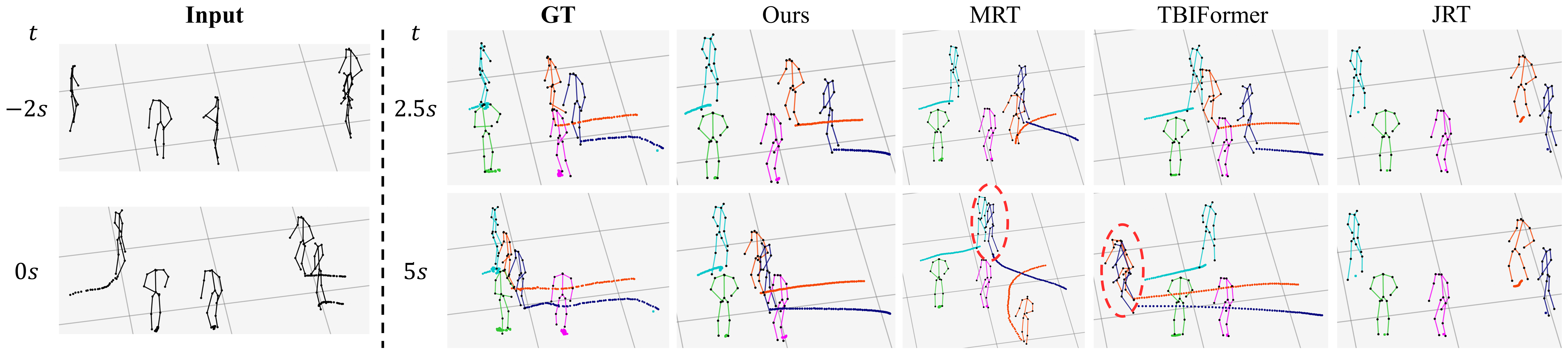}
\vspace{-20pt}
\caption{Visualization of a long-term forecasting scene from JRDB-GMP (2/5) dataset. Past poses for input are shown on the leftmost column, GT future poses on the next, and forecasts by ours, MRT, TBIFormer, and JRT, respectively.}
\label{Results:fig6_jrdb_viz}
\vspace{-15pt}
\end{figure*}
\begin{figure}[t]
\includegraphics[width=\linewidth]{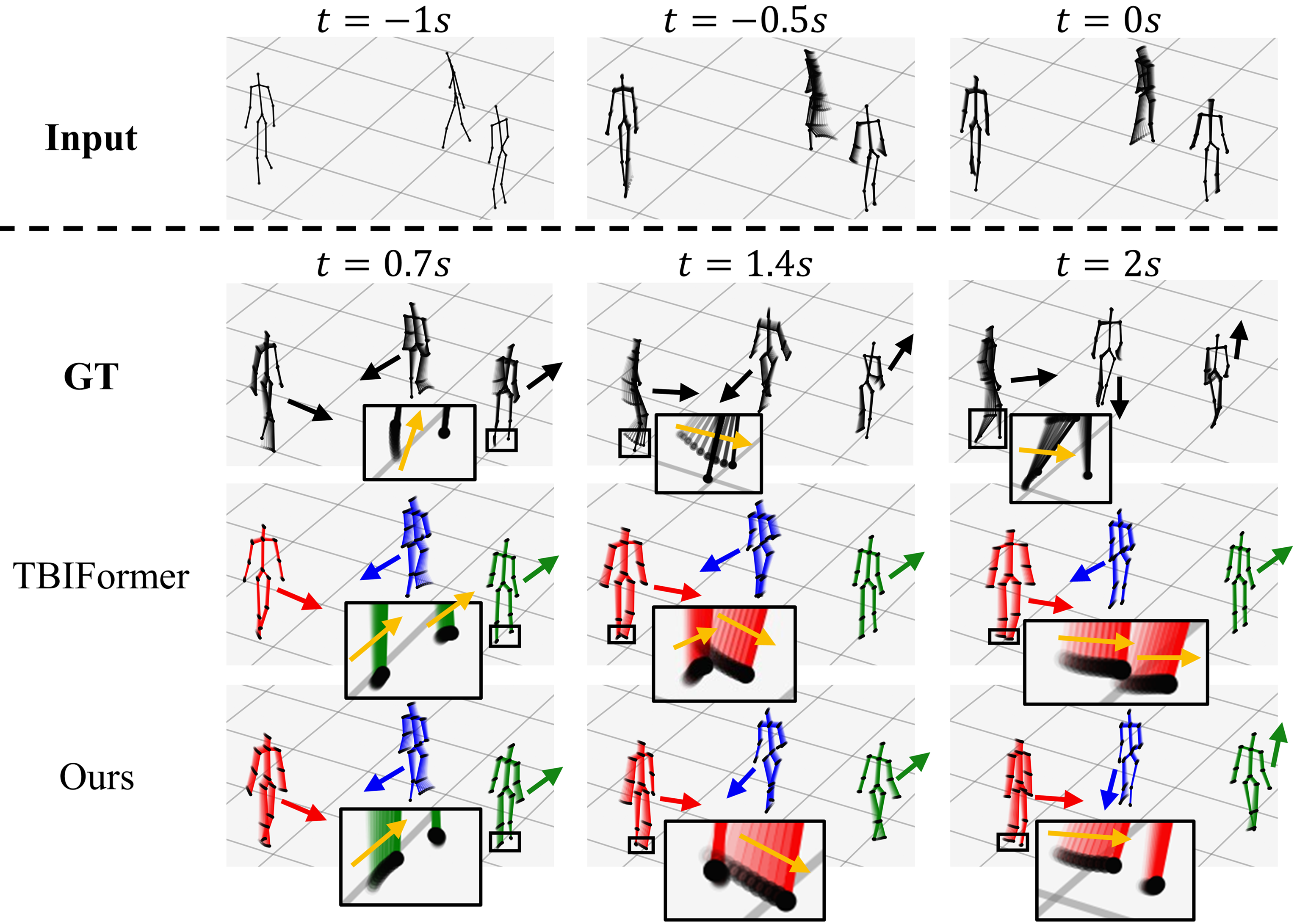}
\vspace{-17pt}
\caption{Visualization of a CMU-Mocap (UMPM) scene. Past poses are shown on the upper row, GT future poses on the next, and forecasts by TBIFormer and ours on the latter two rows.
To visualize motion, we stack several frames around the target time stamp.
Black/red/blue arrows refer to the direction of the global trajectory, and yellow arrows refer to the direction of foot motion. }
\label{Results:fig5_cmu_viz}
\vspace{-14pt}
\end{figure}

\subsection{Baselines}
We compare our method against the latest SOTA methods for multi-agent pose forecasting~\cite{wang2021multi, peng2023trajectory, xu2023joint}.
To compare the multi-modal predictions of these three methods, we extend their prediction modes by spanning embedding $K$ times in the same manner as ours.
All baselines are trained and evaluated on CMU-mocap (UMPM) and 3DPW datasets.
CMU-Mocap (UMPM) dataset predicts 2 seconds from 1 second of poses, and 3DPW predicts 1.6 seconds from 0.8 seconds of poses, both from 6 modes.
For 3DPW dataset, we slightly lengthen the forecast horizon to evaluate long term predictions.
For JRDB-GMP dataset, both short (1s/2s) and long term (2s/5s) predictions are evaluated for all models.
Lastly, We use HiVT~\cite{zhou2022hivt} as the baseline For global trajectory prediction of our model.

\section{Results}
\subsection{Quantitative results}
Table \ref{tab:main_results} compares the quantitative performances on three datasets.
Our method exhibits considerable performance gain against all previous SOTA methods, not only on the proposed long-term multi-agent dataset but also on the existing two datasets.
Such generalized competence demonstrates the applicability of our trajectory-conditioned pose forecasting method to various real-world scenarios.
In detail, our approach achieves over 10\% gain of FDE on all datasets.
This improvement on forecasting global locomotion could be accredited to the decoupled forecasting of global trajectory and local pose.
Previous methods holistically predict both global and local movements, limiting both performances due to superfluous interactions to consider between all joints.
Conversely, our approach can extract accurate global intent by decoupling past motion into global and local representations.
Moreover, our effective interaction modeling of global and local pose also helps to predict a more accurate global trajectory under multi-agent environment as shown in the latter ablation study.

For APE metric, our method also surpasses previous SOTA models on all datasets, highlighting the accurate extraction of local pose intent.
% from global motion via global trajectory-conditioning.
Such improvement shows that our approach generates plausible local motion due to its proficient sampling from coarse global intents. 
% Specifically, multi-modality which means diversity and plausibility, gets more important in long-term prediction.
% Other methods attempt to predict overall motion at once, an approach with excessive complexity that limits adequate modeling of local pose entangled with multi-modality.
Our method simplifies the task by learning multi-modality in a coarse-to-fine approach.
%msimilar to previous works~\cite{salzmann2023robots}.
Its subsequent local motion forecasting is inferred from coarsely modeled multi-modality, a greatly simplified task compared to extracting intent from entangled multi-modality as well as multi-agent interaction.
% However, in our method, which attempt to learn multi-modal nature by conditioning inferred global intent, with the fact that it is much easier to learn multi-modality in coarser scale as following previous works~\cite{salzmann2023robots}.

% that forecasting local motion from global trajectory proposals accurately extracts the local pose intent from global motion intent.
These improvements on both global and local scales jointly contribute toward lowering the JPE metric, demonstrating proficiency of our method in forecasting overall human motion.
Based on such competence, our method which aimed towards improving on forecasting long-term multi-agent environments also exhibits similar or better performances on short timeframes as shown in Tab.~\ref{tab:ablation_short_term}.
In addition, our approach also excels even on sole local motion with minimal global displacement, as elaborated in section 7.2 of supplementary materials.

% \begin{table}[]
% \centering
% \small
% \caption{Comparsion of performance with different number of modes in CMU-Mocap (UMPM) dataset.}
% \label{tab:ablation_num_modes}
% \resizebox{\linewidth}{!}{
% \begin{tabular}{l|ccc|ccc}
% \toprule
% \makecell{$F$} & \multicolumn{3}{c|}{1} & \multicolumn{3}{c}{6} \\ \hline
% \makecell{Metric @ 2s}                & APE    & JPE   & FDE   & APE   & JPE   & FDE   \\ \hline
% MRT             & 163.9       & \textbf{366.4}      & 313.3      & 164.4      & 280.1      & 204.7      \\
% JRT             & 176.7       & 367.4      & \textbf{305.0}      & 181.6      & 316.9      & 250.8      \\
% TBIFormer             & 160.1       & 374.3      & 316.3      & 160.8      & 290.9      & 228.5      \\
% Ours                  & \textbf{154.4}       & \textbf{366.4}      & \textbf{306.5}      & \textbf{151.7}      & \textbf{262.7}      & \textbf{188.9}      \\ \bottomrule
% \end{tabular}
% }
% % \vspace{-15pt}
% \end{table}
\subsection{Qualitative results}
% In addition to such proficiency in forecasting detailed local motion in short timescales ($\sim$1s), o
Our method forecasts a more plausible global pose in longer timescales ($\sim$5s) as shown in the interacting scene of five agents in Fig.~\ref{Results:fig6_jrdb_viz}.
Looking at the input and GT sequences, the leftmost person avoids the traversing couple from right to left.
The two people in front are stationary while talking to each other.
MRT and TBIFormer forecasts implausible overlapped poses at the final prediction horizon (t=5s).
JRT fails to learn the global locomotion of agents due to the high complexity of its attention mechanism and is stuck in the local minimum of predicting the inactivity of all agents.
On the other hand, our model forecasts plausible poses where the closely interacting two agents walk side-by-side.
% Overall, our method results in more plausible forecasts both in short and long-term observation and also in both local and global coordinates compared to previous methods.

Figure \ref{Results:fig5_cmu_viz} illustrates exemplary sequences where more natural local motion has been forecasted by our method.
Comparing forecasts on a scene of walking agents, our method generates a much more plausible sequence where the stepping foot remains stationary.
On the other hand, the previous SOTA method, TBIFormer, struggles to learn the natural walking mechanism of human legs and a parallel translation of both feet is exhibited.
Such discrepancy shows that trajectory-conditioning for inferring local motion from global intent generates more proficient details in human motion than SOTA methods.
More visualizations could be found in the supplementary materials.

\subsection{Ablation studies}
\begin{table}[t]
\centering
\small
\caption{Comparsion of performance with different number of modes in CMU-Mocap (UMPM) dataset.}
\vspace{-5pt}
\label{tab:ablation_num_modes}
% \resizebox{\linewidth}{!}{
\begin{tabular}{l|cc|cc}
\toprule
\makecell{$F$} & \multicolumn{2}{c|}{1} & \multicolumn{2}{c}{6} \\ \hline
\makecell{Metric @ 2s}                & APE    & JPE     & APE   & JPE      \\ \hline
MRT             & 163.9       & \textbf{366.4}           & 164.4      & 280.1            \\
JRT             & 176.7       & 367.4            & 181.6      & 316.9            \\
TBIFormer             & 160.1       & 374.3           & 160.8      & 290.9          \\
\rowcolor[rgb]{0.9,0.9,0.9} Ours                  & \textbf{154.4}       & \textbf{366.4}           & \textbf{151.7}      & \textbf{262.7}          \\ \bottomrule
\end{tabular}
% }
\vspace{-5pt}
\end{table}
\begin{table}[t]
\caption{Ablation studies on core components of model structures. Experiments are done with JRDB-GMP dataset to evaluate multi-agent long-term performance.}
\vspace{-5pt}
\label{tab:ablation_components}
\resizebox{\linewidth}{!}{
\begin{tabular}{c|ccc|ccc}
\toprule
\multirow{2}{*}{\begin{tabular}[c]{@{}c@{}}Exp. \\ \#\end{tabular}} & \multicolumn{2}{c|}{Trajectory encoder}                                                                                                                         & Pose decoder                                                        & \multicolumn{3}{c|}{Metrics}                                                                  \\ \cline{2-7} 
                        & \multicolumn{1}{c|}{\begin{tabular}[c]{@{}c@{}} Local pose \\ embedding\end{tabular}} & \multicolumn{1}{c|}{\begin{tabular}[c]{@{}c@{}}Agent \\ interaction\end{tabular}} & \begin{tabular}[c]{@{}c@{}} Trajectory\\ -conditioning\end{tabular} & \multicolumn{1}{c|}{\begin{tabular}[c]{@{}c@{}}JPE \\ @5s\end{tabular}} & \multicolumn{1}{c|}{\begin{tabular}[c]{@{}c@{}}APE \\ @5s\end{tabular}} & \multicolumn{1}{c|}{\begin{tabular}[c]{@{}c@{}}FDE \\ @5s\end{tabular}} \\ \hline
-                       &                                                                                     &                                                                                      &                                                                       & 471.4                         & 101.7                         & 457.9                         \\ 
1                       &                                                                                     & \checkmark                                                                                    &                                                                       & 400.5                         & 95.1                          & 370.9                         \\ 
2                       & \checkmark                                                                                   &                                                                                      &                                                                       & 403.3                         & 94.7                          & 374.2                         \\ 
3                       &                                                                                     &                                                                                      & \checkmark                                                                     & 401.2                         & 93.0                          & 372.8                         \\
4                       & \checkmark                                                                                   &                                                                                      & \checkmark                                                                     & 395.6                         & 93.8                          & 366.8                         \\ 
5                       & \checkmark                                                                                   & \checkmark                                                                                    &                                                                       & 392.7                         & 95.2                          & 363.4                         \\ 
6                       & \checkmark                                                                                   & \checkmark                                                                                    & \checkmark                                                                     & \textbf{391.2}                         & \textbf{91.4}                          & \textbf{363.3}                         \\ \bottomrule
\end{tabular}%
}
\vspace{-13pt}
\end{table}

\noindent\textbf{Different number of modes.}
The main quantitative results report prediction results with $F$ as 6 to compare the ability to address the multi-modal nature of human motion during pose forecasting.
Table.~\ref{tab:ablation_num_modes} additionally compares forecast results with $F$ as 1.
Our method again achieves noticeable improvement in APE over the baseline on single-modal forecasts.
With $F=1$, although our method barely enjoys improvement in forecasting global motion due to the absence of multi-modality, its superiority in APE shows the validity of our coarse-to-fine forecasting strategy that also effectively captures agent interaction.
% Our method again achieves noticeable improvement over previous methods on single-modal forecasts.
Our method improves with multi-modal predictions, demonstrating the proficiency of a coarse-to-fine approach in interpreting the stochastic nature of human motion and its intent.
Note that our method improves in APE along with an increase in $F$ unlike previous methods, indicating a unique aptitude in addressing the multi-modal nature of not only global locomotion but also local pose intent via trajectory-conditioning.
% Note that our method improves in APE metric along with an increase in $K$, indicating a unique aptitude in interpreting local pose intent via multi-modal global prediction-conditioning.\\

\noindent\textbf{Importance of each architecture component.}
Table.~\ref{tab:ablation_components} reports the influence of core components of our model.
For the trajectory encoder, we evaluate the importance of using local pose embedding and modeling agent interaction.
Comparing experiments 1, 5 and 3, 4, both show improvements in JPE and FDE metrics with the use of local pose embedding.
Our method has taken advantage of detailed local pose cues to infer an agent's global intention.
% Note that its degree of improvement in JPE and FDE is greater with agent interaction (exp. 1,5).
% Indeed, considering the interaction between local motion cues amplifies the benefits of inferring global motion intents for all agents.
For interaction modeling, its use is beneficial for both global and local forecasts as compared by experiments 4 and 6.
These joint improvements demonstrate the importance of considering local and global motion interactions for their respective forecasts.
As for the pose decoder, comparisons of experiments 2,4 and 5,6 both show improvements in APE metric.
Such consistent improvement verifies the effectiveness of the trajectory-conditioned local motion forecast approach in generating plausible local motion from global intention.
% Supplementary materials for more ablations on CMU-Mocap (UMPM) dataset.
% Comparing experiments 2, 5 and 4, 6, modeling agent interaction allows the model to predict global locomotion with the global movement of surrounding agents in consideration. and move accordingly.
% We consider three factors to infer intent on future locomotion: past global trajectory, past local motion, and agent interaction.
% First, use of motion embedding 
% Table \ref{tab:ablation_components} investigates the impact of each factor on forecasting future global motion.
% Matching common intuition, the model performance starkly degrades without knowledge of past trajectory.
% Inter-agent interaction also significantly contributes toward accurate forecasting of global trajectory, improving 4.5\% when the multi-modal interaction is considered.\\

\noindent\textbf{Importance of interaction modeling.}
Accurate modeling of inter-agent interaction becomes more pivotal to forecast in more complex environments.
Indeed, its complexity grows in a long-term multi-agent scene.
%, requiring accurate yet efficient interaction modeling.
When holistically considering joint-wise interaction for all timesteps, the computation complexity is acquired as $O(T^2\cdot N^2\cdot J^2)$, where $T$ is the number of timesteps, $N$ the number of agents, and $J$ the number of joints.
On the contrary, with interaction modeling in global trajectory scale, our method reduces the computation cost by $TJ^2$ into $O(T\cdot N^2)$.
This enables efficient and proficient modeling of intra (pose) and inter (trajectory)-agent interactions as shown by Tab.~\ref{tab:Ablation_interaction}.
While body part-wise interaction modeling only improved by 0.52\% for TBIFormer, ours improves up to 3.84\% with interaction modeling.
% Our interaction modeling method contributes to the significant improvement of all metrics on the complex scenes of JRDB-GMP dataset.
This demonstrates the proficiency of our efficient interaction modeling-based method in inferring global and local intents from complex interactions.
In addition, the gradual improvement of JPE according to a wider interaction range confirms the importance of interaction modeling of more agents, which cannot be learned from the arbitrarily mixed previous datasets.

\begin{table}[]
\caption{Ablation studies on agent interaction cutoff distance on JRDB-GMP.}
\vspace{-5pt}
\label{tab:Ablation_interaction}
\centering
\small
\begin{tabular}{l|cc}
\toprule
                             & \multicolumn{2}{c}{JPE @ 5s} \\ \cline{2-3} 
                             & TBIFormer       & Ours       \\ \hline
w/o interaction              & 483.8           & 406.0      \\
w/ interaction \textless{} 2m  & -               & 403.5      \\
w/ interaction \textless{} 4m & -               & 400.5      \\
w/ interaction all           & 481.3           & \textbf{390.4}     \\ \bottomrule
\end{tabular}
\vspace{-13pt}
\end{table}
\section{Conclusion}

% GPT 버젼
In this work, we propose a novel interaction-aware trajectory-conditioned approach to handle long-term multi-agent motion forecasting, along with a new dataset suited for such scope.
Our proposed model utilizes a coarse-to-fine approach and decouples overall motion prediction into global and local components.
Multi-modality of human motion is proficiently modeled via inferring fine local intents from coarse global intents, along with efficient agent-wise interaction modeling.
As for the dataset, our JRDB-GMP dataset contains unprecedented long-term (5s+) multi-agent (6+) interactions in a real-world setting.
Our method achieves state-of-the-art performance on all previous datasets and JRDB-GMP dataset, offering generalized practical implications in real-world applications.
\newline\noindent\textbf{Acknowledgements}
This research was supported by National Research Foundation of Korea (NRF) grant funded by the Korea government (MSIT) (NRF2022R1A2B5B03002636) and the Challengeable Future Defense Technology Research and Development Program through the Agency For Defense Development (ADD) funded by the Defense Acquisition Program Administration (DAPA) in 2024 (No.912768601).
% This work was supported by Institute of Information \& Communications Technology Planning \& Evaluation(IITP) grant funded by Korea government(MSIT) (No.2019-0-01126, Self-learning based Autonomic IoT Edge Computing). (50\%) and the Korean National Police Agency. [Project Name: XR Counter-Terrorism Education and Training Test Bed Establishment / Project Number: PR08-04-000-21] (50\%)
% addresses the limitations of existing long-term multi-agent human pose forecasting methods by introducing a novel global-trajectory conditioned prediction model. 
% The proposed model efficiently separates motion prediction into global and local components, utilizing the coarse-to-fine information encoded in a person's global trajectory, leading to state-of-the-art performance. 
% By overcoming challenges in predicting beyond short time horizons and modeling interactions among multiple agents, this research significantly advances the capabilities of human pose forecasting models, with potential applications in autonomous driving, robot planning, and surveillance.
% Furthermore, the model's generalizability is demonstrated by achieving state-of-the-art performance not only on the introduced dataset but also on existing datasets. 
% Our research contributes valuable insights and advancements to the field of long-term multi-agent human pose forecasting, offering practical implications for real-world applications.

{
    \small
    \bibliographystyle{ieeenat_fullname}
    \bibliography{main}
}


\end{document}